\documentclass[journal=jctc,manuscript=article]{achemso}
\usepackage{chemformula} % Formula subscripts using \ch{}
\ProvideTextCommand{\DJ}{OT1}{\leavevmode\raisebox{-.5ex}{\makebox[0pt][l]{\hskip-.07em\accent"16\hss}}D}
\usepackage{adjustbox}
\usepackage{caption}
\usepackage{subcaption}
\usepackage{color}
\usepackage{hyperref}
\usepackage{booktabs}
\usepackage{multirow}

\usepackage[flushleft]{threeparttable}
\usepackage{booktabs,caption}
\usepackage{amsmath}
\usepackage{makecell}

\author{Tianhang Zhou}
\affiliation[CUPB]{State Key Laboratory of Heavy Oil Processing, College of Carbon Neutrality Future Technology, China University of Petroleum (Beijing), Beijing 102249, China}
\email{zhouth@cup.edu.cn}
\author{Yingchun Niu}
\affiliation[CUPB]{State Key Laboratory of Heavy Oil Processing, College of Carbon Neutrality Future Technology, China University of Petroleum (Beijing), Beijing 102249, China}
\author{Xingying Lan}
\affiliation[CUPB]{State Key Laboratory of Heavy Oil Processing, College of Carbon Neutrality Future Technology, China University of Petroleum (Beijing), Beijing 102249, China}
\author{Chunming Xu}
\affiliation[CUPB]{State Key Laboratory of Heavy Oil Processing, College of Carbon Neutrality Future Technology, China University of Petroleum (Beijing), Beijing 102249, China}

\title[An \textsf{achemso} demo]
  {Locally-Deployed Chain-of-Thought (CoT) Reasoning Model in Chemical Engineering: Starting from 30 Experimental Data}

\abbreviations{IR,NMR,UV}

\DeclareUnicodeCharacter{2212}{-}

\begin{document}
\pagebreak
\begin{tocentry}

\includegraphics[scale=0.25]{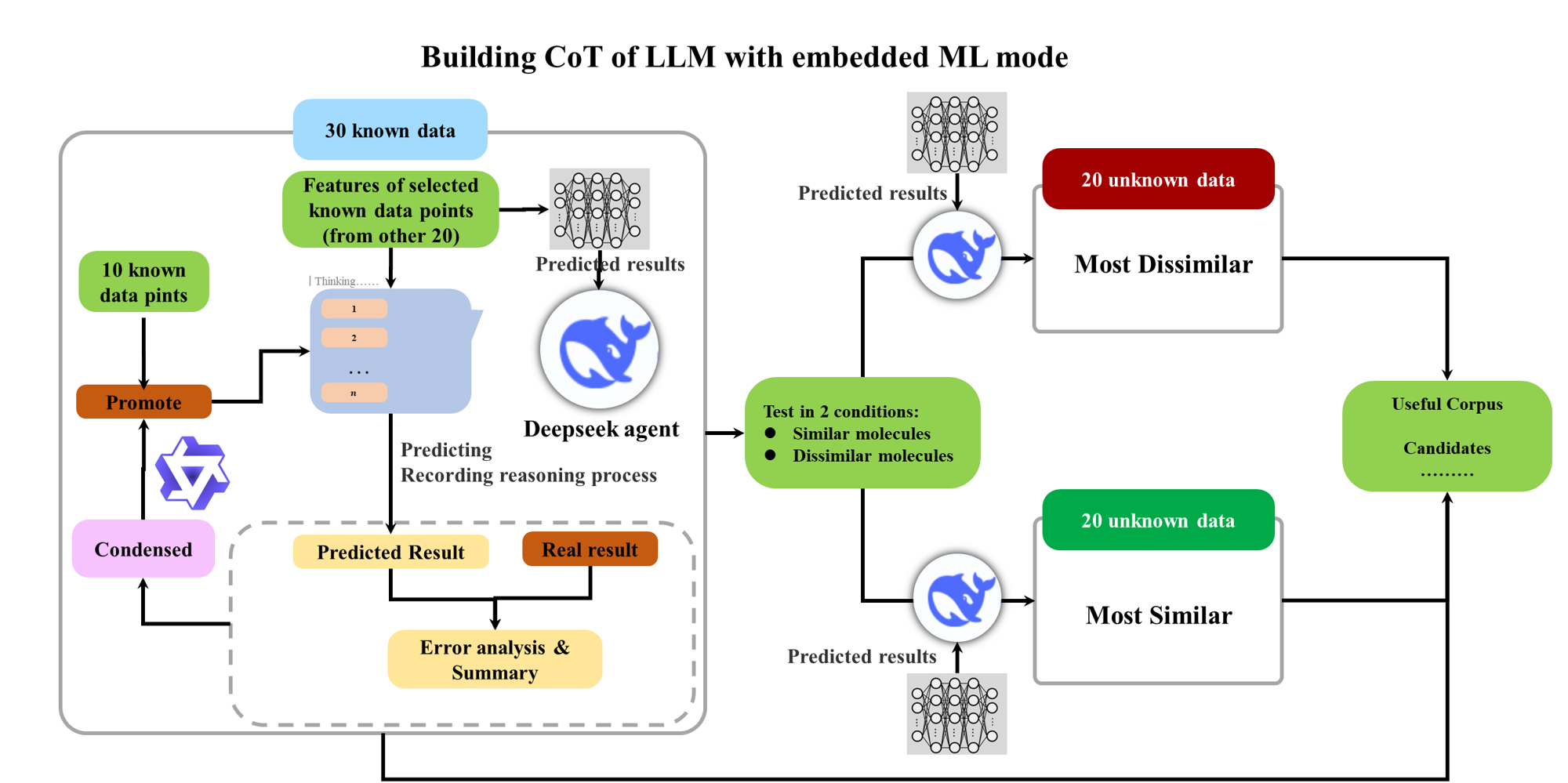}

\end{tocentry}

\begin{abstract} 
In the field of chemical engineering, traditional data-processing and prediction methods face significant challenges. Machine-learning and large-language models (LLMs) also have their respective limitations. This paper explores the application of the Chain-of-Thought (CoT) reasoning model in chemical engineering, starting from 30 experimental data points. By integrating traditional surrogate models like Gaussian processes and random forests with powerful LLMs such as DeepSeek-R1, a hierarchical architecture is proposed.
Two CoT-building methods, Large Language Model-Chain of Thought (LLM-CoT) and Machine Learning-Large Language Model-Chain of Thought (ML-LLM-CoT), are studied. The LLM-CoT combines local models DeepSeek-r1:14b and Qwen2:7b with Ollama. The ML-LLM-CoT integrates a pre-trained Gaussian ML model with the LLM-based CoT framework.
Our results show that during construction, ML-LLM-CoT is more efficient. It only has 2 points that require rethink and a total of 4 rethink times, while LLM-CoT has 5 points that need to be re-thought and 34 total rethink times. In predicting the solubility of 20 molecules with dissimilar structures, the number of molecules with a prediction deviation higher than 100\% for the Gaussian model, LLM-CoT, and ML-LLM-CoT is 7, 6, and 4 respectively. For molecules with similar structures, the Gaussian model and ML-LLM-CoT both have 0 molecules with a prediction deviation higher than 100\%, while LLM-CoT has 3. In terms of solubility judgment success count for 20 dissimilar molecules, the Gaussian model has 15, LLM-CoT has 16, and ML-LLM-CoT has 18. For 20 similar molecules, the Gaussian model and ML-LLM-CoT both have a success count of 20, while LLM-CoT has 17.
These results indicate that ML-LLM-CoT performs better in controlling the number of high-deviation molecules, optimizing the average deviation, and achieving a higher success rate in solubility judgment, providing a more reliable method for chemical engineering and molecular property prediction. This study breaks through the limitations of traditional methods and offers new solutions for rapid property prediction and process optimization in chemical engineering.
\end{abstract}

\pagebreak



\maketitle

\section{Introduction}\label{sec1}

In the realm of chemical engineering, the traditional experimental paradigm for data processing and prediction has long been the cornerstone of research. However, traditional experimental methods often rely on time-consuming and resource-intensive laboratory procedures. For instance, in determining the solubility of molecules, extensive experimental setups are required to measure and analyze the properties under various conditions. These methods typically involve manual data collection and analysis, which are prone to human error and are inherently slow in terms of data-generation speed. Moreover, traditional data-processing techniques for prediction, such as simple statistical models, struggle to handle complex molecular structures and relationships. They often fail to capture the intricate non-linear interactions between different molecular features, leading to inaccurate predictions, especially when dealing with novel or less-studied molecules.

The advent of machine-learning techniques has brought about a significant shift in the field. Machine-learning models, with their ability to learn from large datasets, have shown great promise in molecular-property prediction. For example, models like neural networks can automatically extract features from molecular descriptors, enabling more accurate predictions compared to traditional methods. They are capable of handling high-dimensional data and identifying complex patterns within the data. However, machine-learning also has its own set of limitations. One major drawback is the requirement for a large amount of labeled data for training. In the context of molecular research, obtaining a sufficient number of accurately labeled molecular datasets can be extremely challenging due to the high cost and time-consuming nature of experimental validations. Additionally, machine-learning models often lack interpretability. It can be difficult to understand how these models arrive at their predictions, which is a crucial aspect in scientific research, especially in chemical engineering where understanding the underlying mechanisms is essential.

The introduction of large-language models (LLMs) without reasoning capabilities has also had an impact on the field. These models can process and analyze text-based information related to molecular research, such as chemical literature and experimental reports. They can extract relevant knowledge and information from a vast amount of text data, providing valuable insights for molecular-property prediction. For example, they can help in identifying trends and relationships described in scientific papers that might be overlooked by traditional methods. However, they also have limitations. Without reasoning capabilities, these LLMs are mainly limited to pattern-recognition and text-matching tasks. They cannot perform in-depth causal reasoning about molecular properties and interactions, which restricts their ability to make accurate predictions in complex and novel scenarios.

\begin{figure}[H]
\centering
\includegraphics[width=1.0\textwidth]{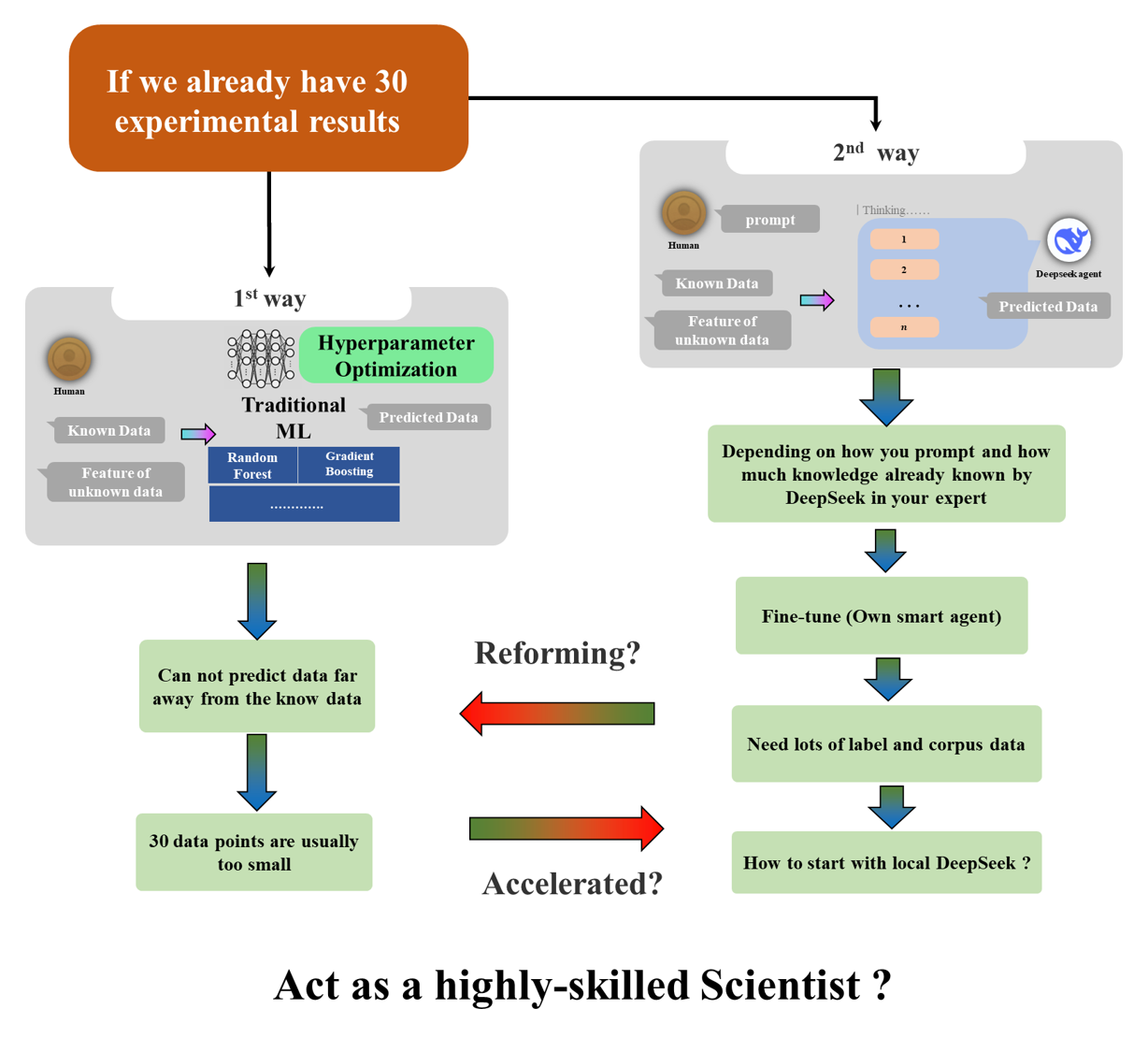}
\caption{A flowchart comparing two approaches for predicting data when 30 experimental results are available: traditional machine-learning with hyperparameter optimization and a method involving the DeepSeek agent, highlighting challenges and considerations such as prediction range, data size, and requirements for fine-tuning.}
\label{fig:figure1}
\end{figure}

Recently, the emergence of DeepSeek\cite{deepseek} (mainly for DeepSeek-R1\cite{DeepSeek2023R1}), a large-language model with reasoning capabilities, has brought the potential for transformation in the science community. People are all speculating about what roles DeepSeek-R1 can play. DeepSeek-R1, in particular, can extract high-order correlations from sparse data, such as molecular-fragment combination rules. This ability is crucial as it allows for a more in-depth understanding of molecular structures and their properties. For example, as illustrated in Figure \ref{fig:figure1}, when we already possess 30 experimental results, two distinct approaches can be considered for molecular-property prediction. The first approach involves traditional machine-learning methods such as Random Forest and Gradient Boosting, which are followed by hyperparameter optimization. However, this traditional route has its limitations. It struggles to predict data that is far from the known data, and a dataset of only 30 data points is usually deemed too small for robust predictions.
In contrast, the second approach utilizes DeepSeek-R1. Here, the known data, along with the features of the unknown data, is inputted as prompts. The effectiveness of the prediction depends on how well the prompts are formulated and the existing knowledge base of DeepSeek-R1 in the relevant expertise. Subsequently, fine-tuning with one's own smart agent may be required, which necessitates a substantial amount of labeled and corpus data. Additionally, the question of how to initiate the use of a local DeepSeek-R1 model also arises. 

Our research team believes that in the fields of chemical engineering, traditional surrogate models like Gaussian processes and random forests possess their own advantages\cite{doi:10.1021/acs.jctc.1c00134,doi:10.1021/acs.jctc.3c01348,doi:10.1021/acs.macromol.2c00821}. Their forte lies in uncertainty quantification with small samples. When data is scarce, they can effectively assess the potential variability in predictions, providing a measure of reliability that is crucial in decision-making processes. For example, in early-stage molecular design projects where only a limited number of experimental results are available, these models can help researchers understand the possible range of outcomes.
On the other hand, DeepSeek-R1, a powerful large-language model, shines in extracting high-order associations from sparse data. It can uncover complex relationships such as molecular fragment combination rules that might be overlooked by traditional methods. This ability to dig deeper into the data fabric enables it to offer unique insights that can drive innovation in molecular prediction.
When these two types of models are used in combination, we envision a hierarchical architecture. The surrogate models take on the task of rapidly screening candidate regions. They can quickly narrow down the search space based on existing data patterns, saving valuable computational time and resources. Meanwhile, DeepSeek-R1 focuses on what we term 'boundary samples'. These are samples with high uncertainty but also the potential for high rewards. By performing causal reasoning on these samples, DeepSeek-R1 mimics the 'intuitive breakthroughs' of human experts. It can explore uncharted territories in the data, making connections that might not be immediately obvious. We suspect that this division of labor could be more efficient than a simple sequential combination of the two models.

Furthermore, during the data prediction process of DeepSeek-R1, it is noticed that it implicitly engages in different types of scenario-based thinking\cite{li2025codeiocondensingreasoningpatterns}. This inherent thinking process presents a valuable opportunity. We aim to 'extract' this thinking and use it to train a reasoning model, which is essentially the concept of Chain-of-Thought (CoT) in chemical engineering. By leveraging this approach, we hope to further enhance the predictive capabilities and reasoning power in our pursuit of more accurate and intelligent solutions in chemical engineering and molecular prediction.

%In this work, we first demonstrate the marked disparities between traditional ML and DeepSeek-based prediction approaches through contrastive experiments, then reveal the adaptability of the Deepseek agent in handling different data similarities by a systematic data-handling process, ultimately establishing a comprehensive workflow that combines different prediction methods for chemical engineering applications. We further propose a comprehensive workflow that combines different prediction methods, which can be applied to a wide range of chemical engineering scenarios. This discovery not only breaks through the traditional limitation of relying solely on machine-learning with scarce data but also provides a new solution path for rapid property prediction and process optimization in chemical engineering.

\section{Building CoT with LLM themself}
\subsection{Preparation and Method}

In this study, we have carefully designed a deployment logic that combines multiple models locally with Ollama\cite{ollama}:
(1) DeepSeek-r1:14b\cite{deepseek} (4.7 GB) offers a balance between computational efficiency and performance. It is deployed for tasks that require relatively quick responses while still maintaining a certain level of accuracy;
(2) Qwen2:7b\cite{qwen} (4.4 GB) is integrated into our framework for its unique strengths in handling diverse language-related tasks in the context of molecular research. It can process chemical literature, patents, and experimental reports efficiently.
Of course, we could optimize results for larger models, such as using the API of DeepSeek-r1:671b. However, the objective of this article is to explore the possibilities of local deployment. For users who possess sensitive data and currently do not have access to a sufficient number of high-performance computers, we attempt to optimize through the construction of Chain-of-Thought. This exploration provides a priori possibilities for subsequently using high-performance supercomputer clusters to call larger models.

By integrating these models and tools in a coordinated manner (named as LLM-CoT), we aim to create a comprehensive and efficient system for molecular-property prediction. The different models complement each other, with the language models handling the reasoning and knowledge-based aspects, and RDKit\cite{rdkit} providing the necessary molecular-specific data processing and representation capabilities. This multi-model deployment strategy allows us to address the complex and diverse challenges in the field of chemical engineering and molecular-property prediction more effectively.

\begin{figure}[H]
\centering
\includegraphics[width=1.0\textwidth]{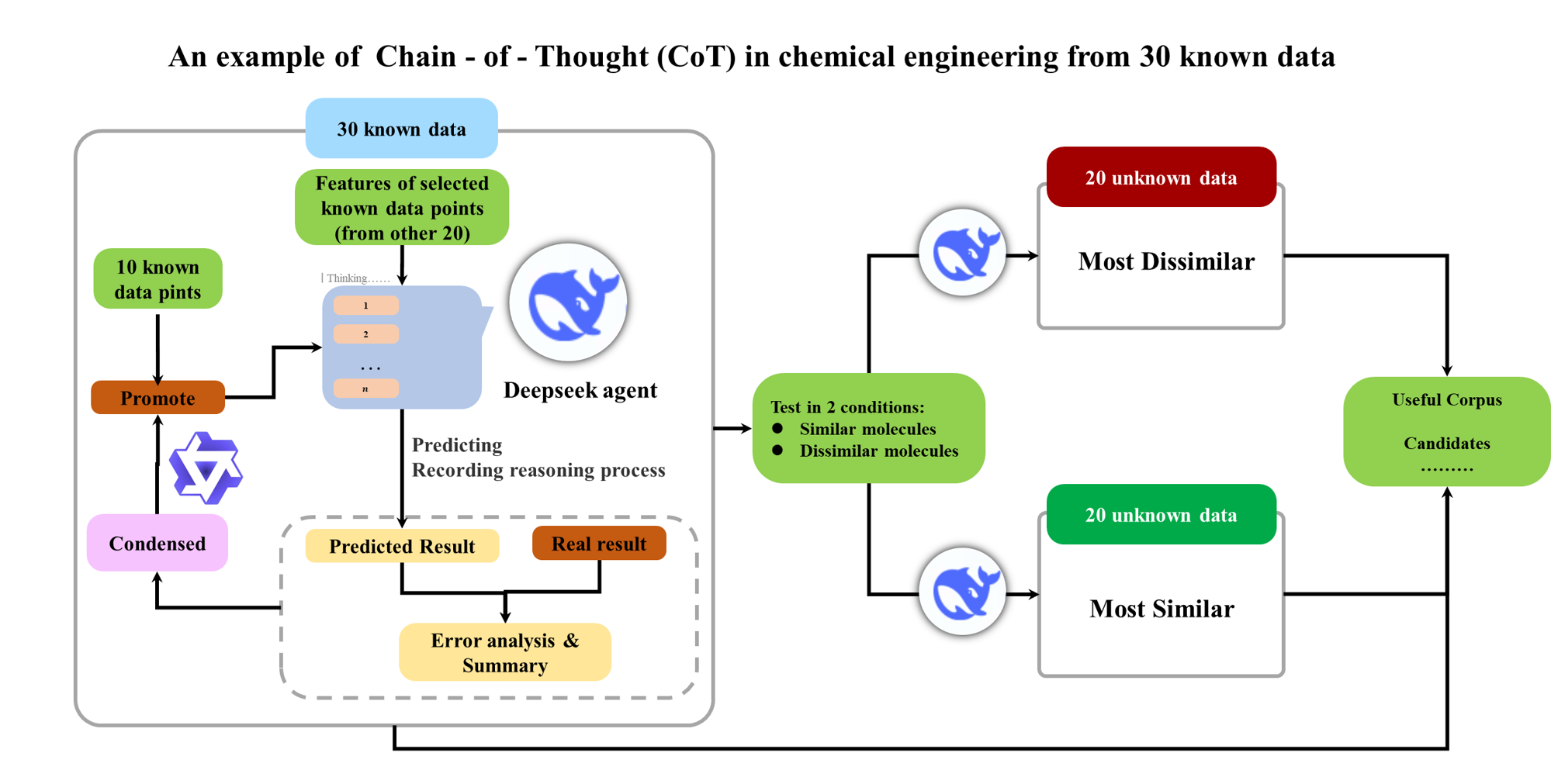}
\caption{Flowchart depicting the process of building a Chain-of-Thought (CoT) using a Deepseek agent. It involves leveraging 30 known data points, promoting and condensing data, predicting with recording of the reasoning process, conducting error analysis and summary, and testing on 20 unknown data points categorized as most similar and most dissimilar, with the aim of creating a useful corpus.}
\label{fig:figure2}
\end{figure}

In regard to data preparation, we first selected the solubility properties of 30 molecules the same 1128-molecule dataset\cite{Subset2020,Wu2017MoleculeNet} along with their molecular descriptors, which could be conveniently retrieved from RDKit. The chosen descriptors included Molecular Weight (MW), LogP, Topological Polar Surface Area (TPSA), Number of Hydrogen Bond Acceptors (NumHAcceptors), Number of Hydrogen Bond Donors (NumHDonors), Number of Rotatable Bonds (NumRotatableBonds), Heavy Atom Count, Number of Aromatic Rings (NumAromaticRings), Fraction of sp3 Hybridized Carbon Atoms (FractionCSP3), and Ring Count. To comprehensively evaluate the prediction performance, we further selected 20 molecules with high similarity and 20 with low similarity to the initial 30 molecules based on molecular similarity metrics from the same dataset. We plan to construct a small cyclic framework specifically tailored to the 30-molecule dataset. In this framework, the DeepSeek-r1:14B model will iteratively make predictions and accumulate error analysis compared to the real data (Figure \ref{fig:figure2})

The prediction analysis starts by loading the initial, least similar, and most similar databases. Variables for error analyses and data storage are initialized. Then, the system loops through the initial database. For each entry, features are calculated, and similar molecules are identified. A prediction prompt is generated based on these calculations. The DeepSeek-R1 model is used to make predictions. After the prediction, the deviation of the prediction is checked. If the deviation is greater than 100$\%$, the system goes back to calculate features again. If the deviation is less than or equal to 100$\%$, Qwen2 is used for error analysis. After the error analysis, the system loops back to the initial database. Once the loop is completed, the system predicts unknown properties based on the previous experience, and the process ends.

\begin{figure}[H]
    \centering
    \includegraphics[width=0.8\textwidth]{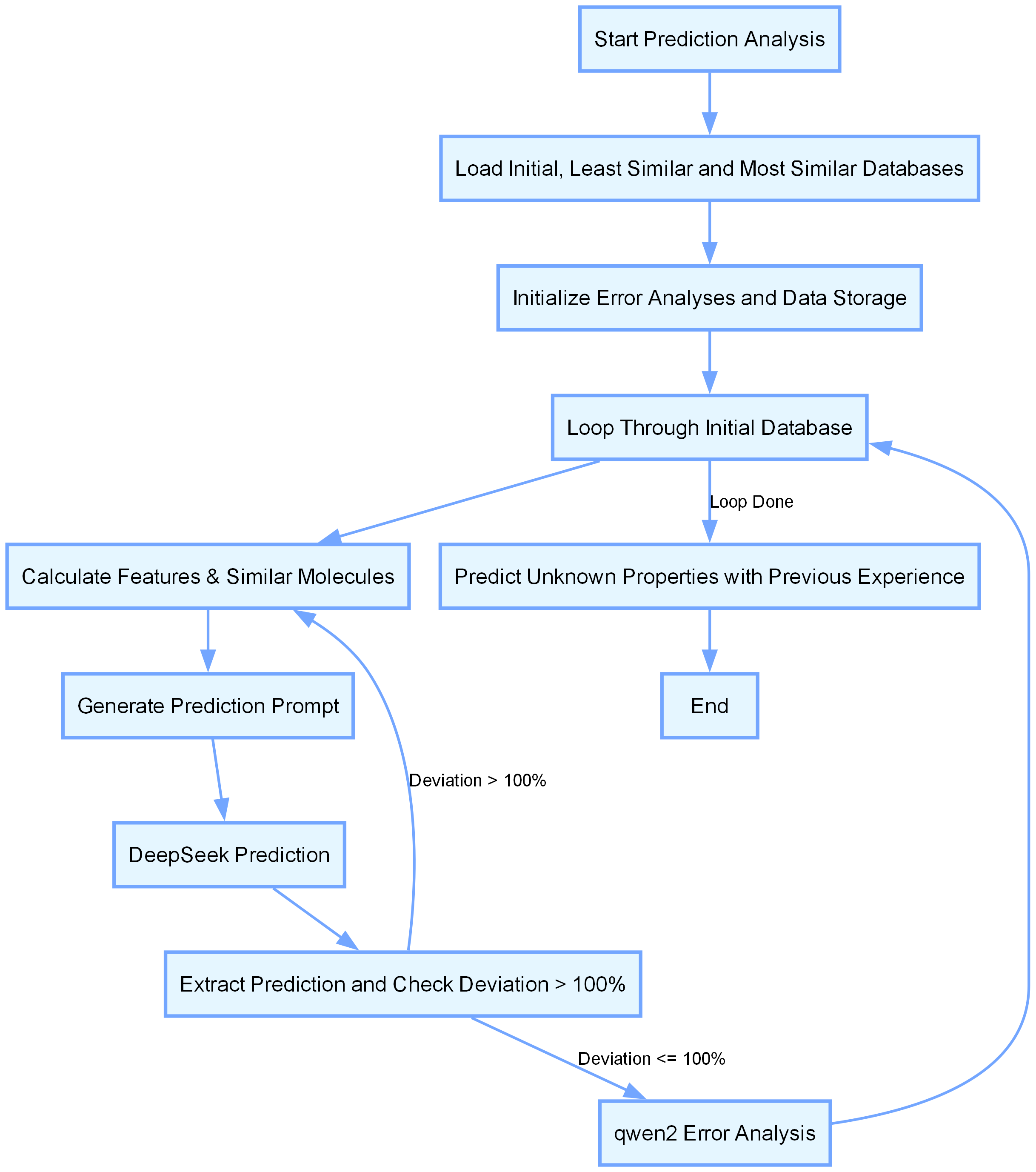}
    \caption{Flowchart of building CoT with LLM themself}
    \label{fig:new_flowchart2}
\end{figure}

\subsection{Result}

Before delving into the results, we made numerous attempts to use the locally-deployed DeepSeek-r1:14b to directly predict 20 unknown data points based on 30 known data points. However, we frequently encountered issues such as missing data points, with the deviation of the predicted values often exceeding even 1000$\%$, or receiving completely irrelevant responses. The root causes of these phenomena lie in the processing capacity limitations of the 14b model and the lack of pre-deployment of relevant content regarding solubility. Paradoxically, this situation provides an excellent experimental ground for validating the Chain-of-Thought  we constructed. It allows us to explore whether the CoT approach can enhance the model's prediction ability for data it was initially unfamiliar with. 

\begin{table}[H]
    \centering
    \begin{threeparttable}
        \caption{Examples of Summarized Unreasonable Responses by directly using DeepSeek-r1:14b}
        \bigskip % 增加垂直间距
        \begin{tabular}{|l|}
        \toprule
            \textbf{Examples of Extremely Unreasonable Responses from DeepSeek-R1:14b} \\
        \midrule
            \makecell[l]{The answer was off-topic.\\[2pt]
            When answering solubility-related questions,\\[2pt]
            it elaborated on the stability of molecules.} \\

            \makecell[l]{The prediction result had a deviation as high as 1000\%.\\[2pt]
            Additionally, the unit of solubility was confused.\\[2pt]
            As a result, the result became meaningless.} \\

            \makecell[l]{Multiple data points were omitted from the prediction.\\[2pt]
            Only some molecules were predicted.} \\
        \bottomrule
        \end{tabular}
        \begin{tablenotes}
            \footnotesize
            \item Note: These unreasonable responses allows us to explore whether the CoT approach can enhance the model's prediction ability for data it was initially unfamiliar with.
        \end{tablenotes}
    \end{threeparttable}
\end{table}

\begin{figure}[H]
    \centering
    \includegraphics[width=0.7\textwidth]{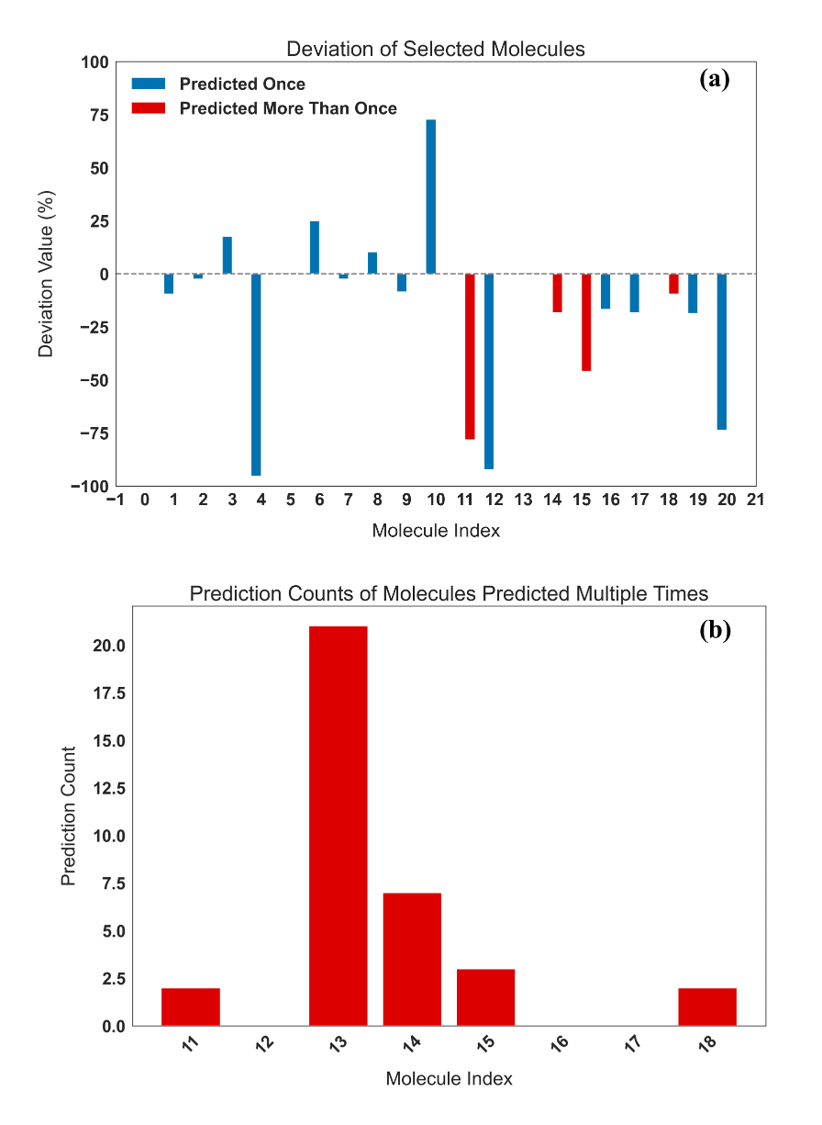}
    \caption{This figure presents the results of the molecular property prediction process during the construction of the Chain-of-Thought (CoT). Sub-figure (a) showcases the single-attempt and multiple-attempt prediction performance of individual molecules, and sub-figure (b) offers in-depth details about the molecules that required multiple predictions, highlighting the importance of error-based iterative prediction and analysis in improving prediction accuracy.}
    \label{fig:llmdiv}
\end{figure}

During the construction of the CoT, the core issue revolved around evaluating the effectiveness of our prediction approach in precisely determining molecular properties. We initiated the process by focusing on the prediction within a set of 30 known molecules. Starting with 30 known data points, we utilized the data of the first 10 molecules as the basis. With the help of structure in Figure \ref{fig:figure2}, we made continuous efforts to predict the properties of the subsequent 20 molecules. During this process, we continuously fed back the prediction results as opinions during the prediction of these 20 molecules. We meticulously monitored and recorded the prediction performance at every step. Notably, if the prediction error was found to be greater than 100$\%$, the prediction cycle would be restarted. This is precisely the reason for the difference between multiple-prediction and single-prediction cases shown in Figure \ref{fig:llmdiv}(a) and Figure \ref{fig:llmdiv}(b). During the process, we observed that some molecules were predicted as many as 20 times. Only after accumulating sufficient error analysis through these multiple attempts could accurate predictions be achieved. We meticulously monitored and recorded the prediction performance at every step. As vividly depicted in Figure \ref{fig:llmdiv}(a), it can be clearly observed that regarding the prediction performance of each molecule, around 60\% of the molecules (depicted by the blue bars) managed to achieve a prediction error of less than 100\% in a single attempt. In contrast, the remaining 40\% (represented by the red bars) necessitated multiple prediction attempts to bring the error rate below 100. Figure \ref{fig:llmdiv}(b) further elaborates by providing detailed information about these molecules that required multiple predictions, which is crucial for us to understand the underlying issues and optimize the CoT-based prediction approach.

\begin{figure}[H]
    \centering
    \includegraphics[width=0.9\textwidth]{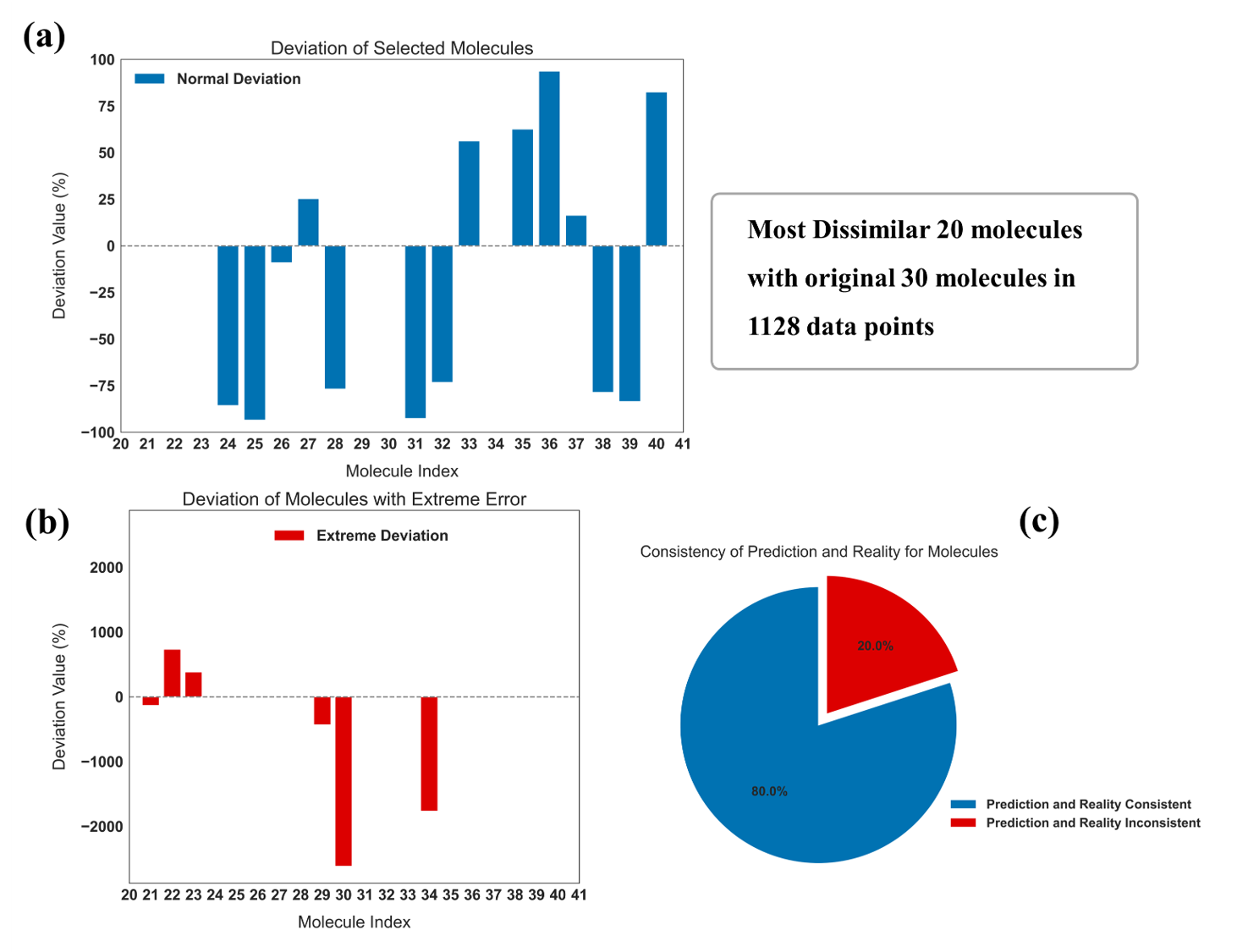}
    \caption{(a) Deviation of the 20 most dissimilar molecules with the original 30 molecules in the 1128-data-point open-source dataset. Blue bars represent normal deviation values of the molecules. (b) Deviation of molecules with extreme error among the 20 most dissimilar molecules. Red bars indicate extreme deviation values. (c) Consistency of solubility determination for the 20 most dissimilar molecules, where the blue part represents the proportion of molecules for which prediction and reality are consistent, and the red part represents the proportion of inconsistent ones.}
    \label{fig:llmdiv1}
\end{figure}

\begin{figure}[H]
    \centering
    \includegraphics[width=0.9\textwidth]{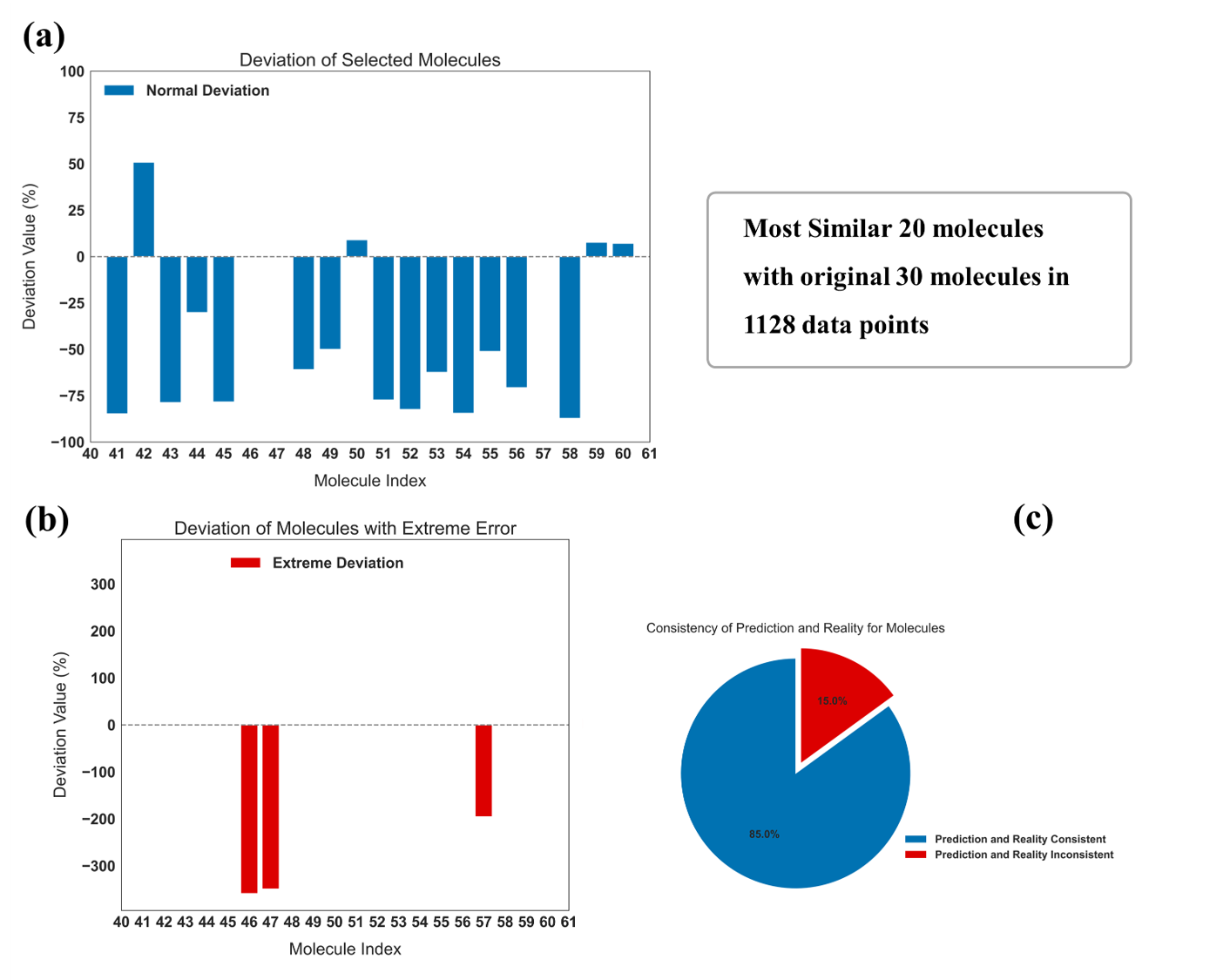}
    \caption{(a) Deviation of the 20 most similar molecules with the original 30 molecules in the 1128-data-point open-source dataset. Blue bars represent normal deviation values of the molecules. (b) Deviation of molecules with extreme error among the 20 most similar molecules. Red bars indicate extreme deviation values. (c) Consistency of solubility determination for the 20 most similar molecules, where the blue part represents the proportion of molecules for which prediction and reality are consistent, and the red part represents the proportion of inconsistent ones. }
    \label{fig:llmdiv2}
\end{figure}

We then proceeded to further test the performance of our constructed Chain-of-Thought  model. We selected 20 molecules with dissimilar structures and 20 molecules with similar structures, all sourced from an open-source dataset containing 1128 data entries, based on molecular similarity compared to the 30 molecules used in the CoT construction process. As illustrated in Figure \ref{fig:llmdiv1} and \ref{fig:llmdiv2}, for the 20 molecules with dissimilar structures, we observed that 6 of them had an error exceeding $100\%$. The remaining 14 molecules had an error lower than $100\%$. For the 20 molecules with similar structures, only 2 had an error higher than $100\%$, while the rest had an error below $100\%$.
In general, the prediction performance for molecules with similar structures was superior to that of dissimilar-structured molecules. There were fewer extreme deviations, and the solubility determination was more accurate. It is important to note that the solubility determination was based on whether the predicted solubility value and the experimental solubility value were both greater than 0 or both less than 0 simultaneously. This new set of experiments not only validates the effectiveness of our CoT-based prediction approach in differentiating between different types of molecular structures but also provides valuable insights for further optimizing the model in the context of chemical engineering and molecular-property prediction.

\section{Building CoT of LLM with embedded ML mode}
\subsection{Preparation and Method}
\begin{figure}[H]
\centering
\includegraphics[width=1.0\textwidth]{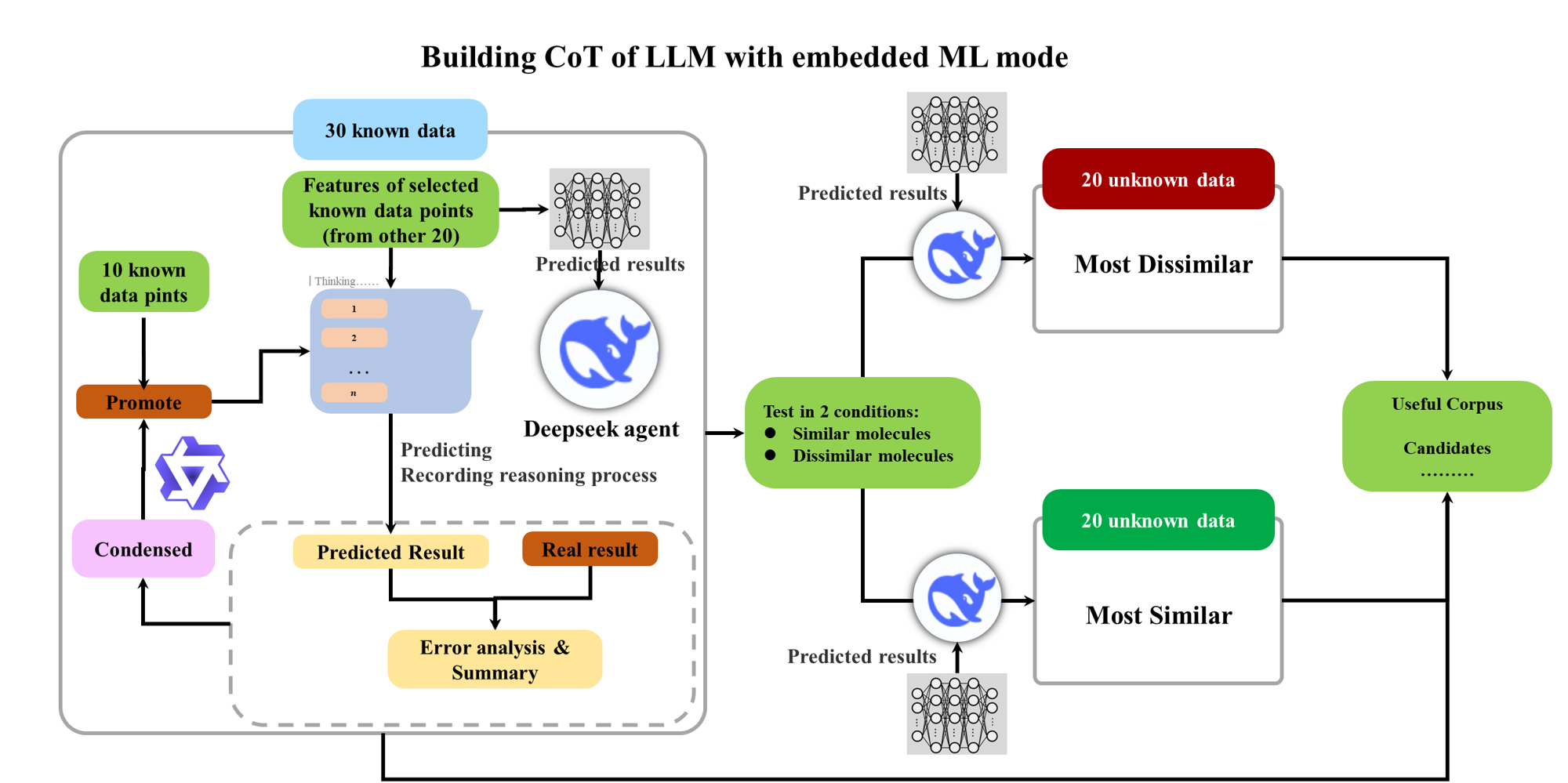}
\caption{The process of integrating Gaussian ML model and LLM for molecular property prediction in the CoT with embedded ML mode. Starting from 30 known data points, Gaussian model makes initial predictions for 20 unknown data points, and then LLM refines these predictions.}
\label{fig:embedded_ml_process}
\end{figure}

\begin{figure}[H]
\centering
\includegraphics[width=0.5\textwidth]{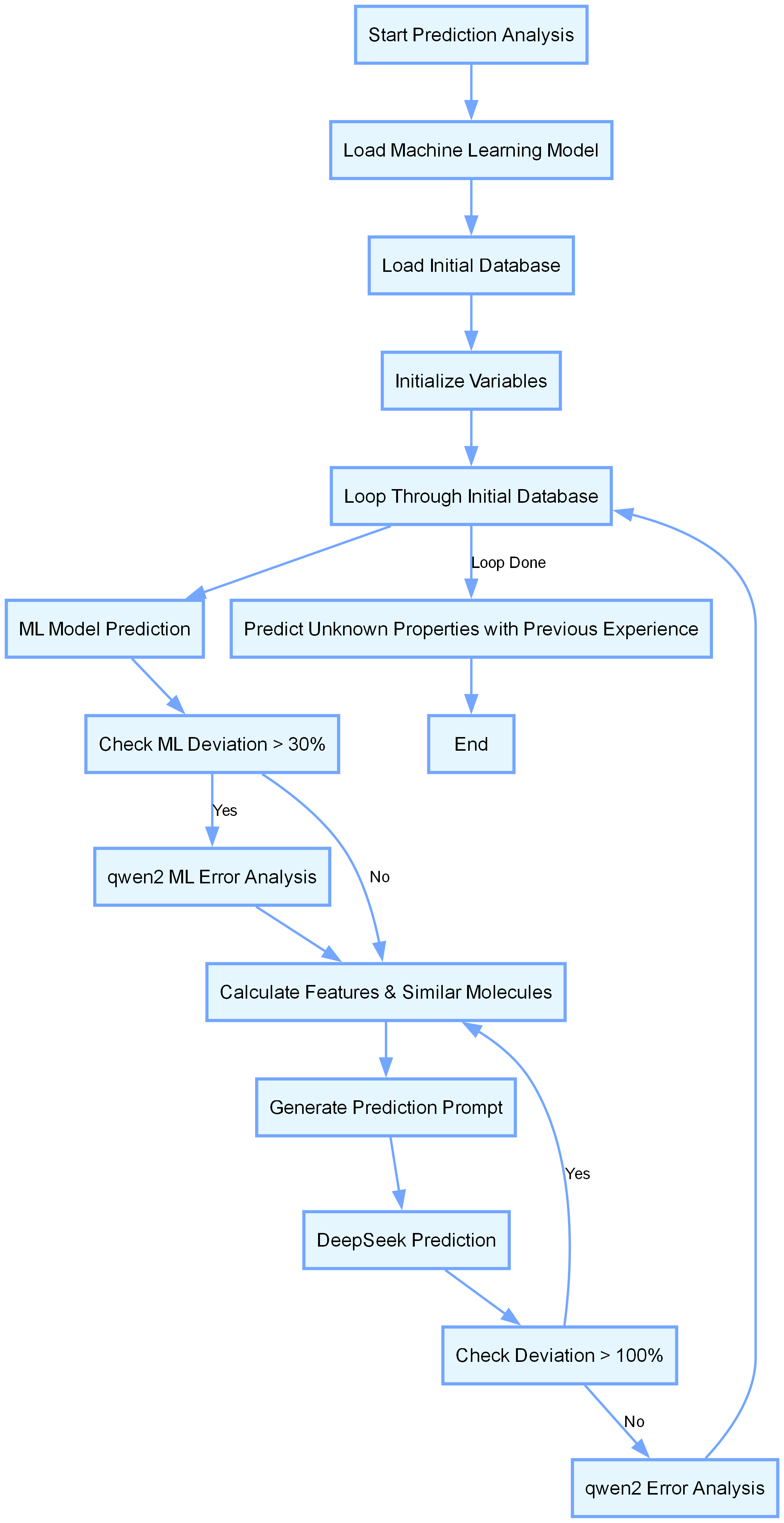}
\caption{Flowchart of the prediction analysis in the CoT with embedded ML mode. It shows the iterative process of using the Gaussian ML model and LLM to make and refine predictions, with error analysis and threshold-based decision-making.}
\label{fig:embedded_ml_flowchart}
\end{figure}

In this approach of building the Chain-of-Thought  with an embedded Machine Learning (ML) mode, we designed an integrated system that combines the strengths of LLM and ML models (named as ML-LLM-CoT). We chose to work with the same 1,128-molecule dataset as in the previous section. From this dataset, we again selected 30 molecules as the basis for our analysis. We used the Gaussian ML model, which we had pre-trained, and integrated it with the LLM-based CoT framework. The Gaussian model was trained to capture the complex relationships between molecular descriptors (such as Molecular Weight (MW), LogP, Topological Polar Surface Area (TPSA), etc., which were retrieved from RDKit) and solubility properties. The overall process is depicted in Figure \ref{fig:embedded_ml_process} and \ref{fig:embedded_ml_flowchart}. In Figure \ref{fig:embedded_ml_process}, we start with the 30 known data points. We first use the Gaussian model to make initial predictions for the 20 unknown data points (both similar and dissimilar to the initial 30 molecules). Then, the LLM (such as DeepSeek-r1:14b or Qwen2:7b) is integrated to refine these predictions. The LLM processes the results from the Gaussian model, along with relevant chemical knowledge and reasoning, to improve the accuracy of the predictions.

Figure \ref{fig:embedded_ml_flowchart} shows the detailed flowchart of the prediction analysis. The system first loads the Gaussian ML model and the initial database. Variables for error analysis and data storage are initialized. Then, it loops through the initial database. For each molecule, the Gaussian model predicts its properties. After that, the deviation of the prediction is calculated. If the deviation is greater than a certain threshold (for example, 30\% as set in our system), the LLM is used to analyze the error and generate a new prediction prompt. The LLM then makes a refined prediction, and the deviation is checked again. This iterative process continues until the deviation meets the required criteria.

\subsection{Result}
We first analyzed the performance of the Gaussian model (trained by 30 known data with optimized hyperparameters) and compared it with the results of LLM-CoT for predicting the solubility of molecules. For the 20 molecules with dissimilar structures, the Gaussian model had 7 molecules with a prediction deviation higher than 100\%. The average deviation for the molecules with a deviation less than 100\% was 30.49, and the average deviation for those with a deviation greater than 100\% was 1253.88. In terms of solubility judgment, it had a success count of 15. For the 20 molecules with similar structures, the Gaussian model performed better. It had 0 molecules with a prediction deviation higher than 100\%, and the average deviation for these molecules was 20.87. The solubility judgment success count was 20. Compared with the LLM-CoT model, for dissimilar molecules, the Gaussian model had more molecules with a high-deviation prediction, but its average deviation for low-deviation molecules was lower. For similar molecules, both models had a relatively high success rate in solubility judgment.

These results demonstrate that the Gaussian model has its own characteristics in predicting molecular solubility. While it struggles with high-deviation cases for dissimilar molecules, it shows great stability for similar-structured molecules. This information is valuable for further optimizing the CoT with embedded ML mode, helping us to better understand how to combine different models to improve the overall prediction performance in chemical engineering and molecular-property prediction. The analysis of the Gaussian model provides a basis for us to understand the performance of molecular prediction. Next, we analyze the prediction results of ML-LLM-CoT.

When predicting 20 molecules with dissimilar structures, the ML-LLM-CoT model demonstrated certain advantages. The model had only 4 molecules with a prediction deviation higher than 100\%, which is less than the 7 of the Gaussian model. For molecules with a prediction deviation lower than 100\%, its average deviation was 37.54125, and the average deviation of molecules with a deviation higher than 100\% was 424.96, both of which were lower than the corresponding deviation values of the Gaussian model. In terms of solubility judgment, the ML-LLM-CoT model had a success count of 18, higher than the 15 of the Gaussian model. When dealing with 20 molecules with similar structures, the ML-LLM-CoT model also performed remarkably. Similar to the Gaussian model, it had no cases where the prediction deviation exceeded 100\%. The average deviation of molecules with a prediction deviation lower than 100\% was 38.49, and the success count of solubility judgment reached 20. Compared with the Gaussian model, when predicting molecules with dissimilar structures, the ML-LLM-CoT model performed better in terms of controlling the number of high-deviation molecules and optimizing the average deviation, and also had a higher success rate in solubility judgment. When predicting molecules with similar structures, both models showed good stability.

\begin{figure}[H]
\centering
\includegraphics[width=0.9\textwidth]{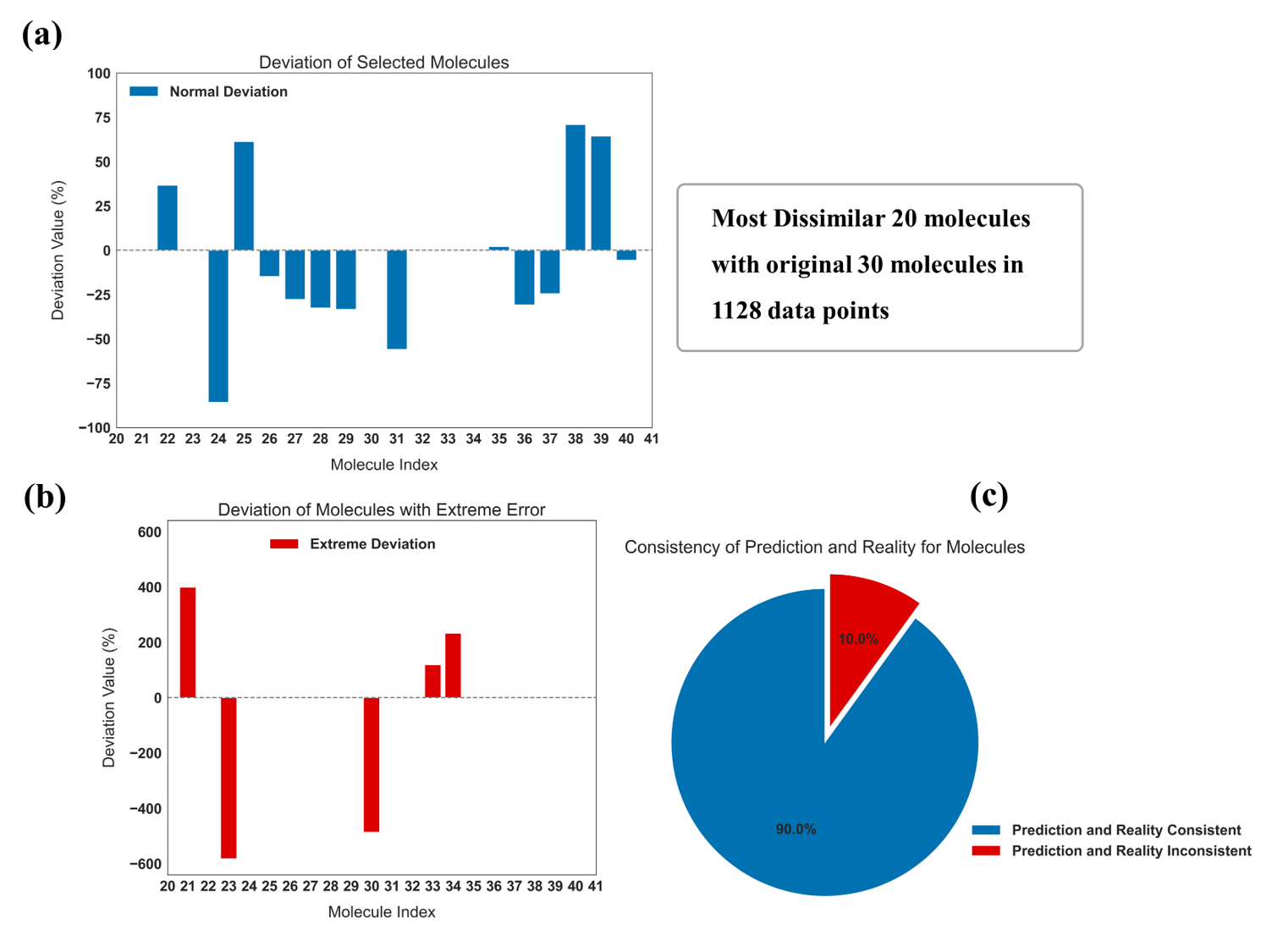}
\caption{(a) Prediction deviation distribution for the 20 most dissimilar molecules using ML-LLM-CoT in the CoT with embedded ML mode. Blue bars represent molecules with a deviation less than 100\%, and red bars represent those with a deviation higher than 100\%. (b) Comparison of average deviations between molecules with low and high deviations among the 20 most dissimilar molecules for ML-LLM-CoT. (c) Solubility judgment success rate for the 20 most dissimilar molecules with ML-LLM-CoT.}
\label{fig:ml_llm_dissimilar}
\end{figure}

\begin{figure}[H]
\centering
\includegraphics[width=0.9\textwidth]{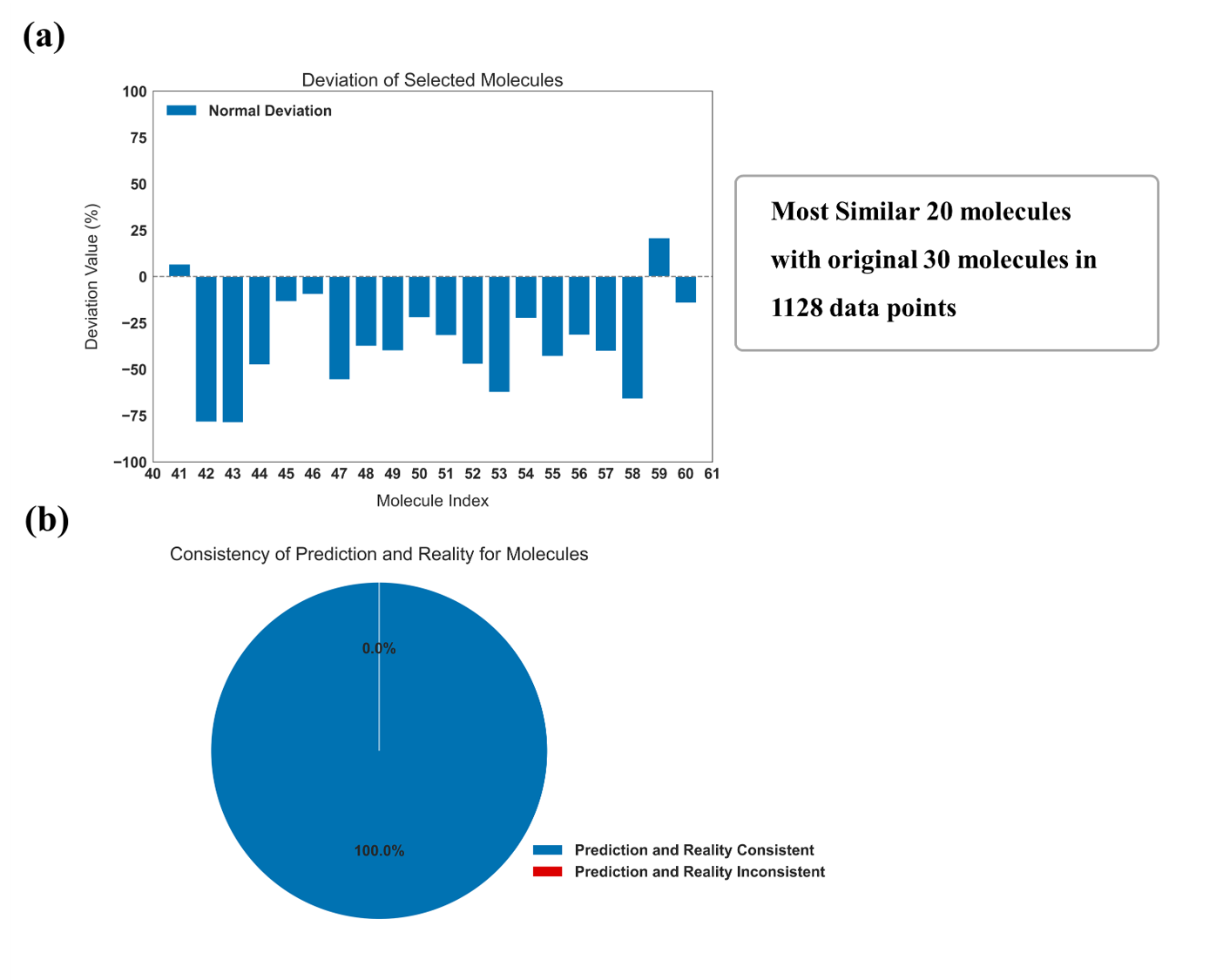}
\caption{(a) Prediction deviation distribution for the 20 most similar molecules using ML-LLM-CoT in the CoT with embedded ML mode. Since there are no molecules with a deviation higher than 100\%, only blue bars representing molecules with a deviation less than 100\% are shown. (b) Solubility judgment success rate for the 20 most similar molecules with ML-LLM-CoT.}
\label{fig:ml_llm_similar}
\end{figure}

\section{Summary and Future Outlook}
\subsection{Summary}
The table \ref{table:rethink_comparison} presents a comparison of the iterative processes of LLM-CoT and ML-LLM-CoT during their construction in terms of the number of points requiring rethink to bring the error rate below 100\% and the total rethink times. For LLM-CoT, there are 5 points that need to be re-thought, and the total number of rethinks amounts to 34. In contrast, ML-LLM-CoT shows a significantly more efficient construction process. It only has 2 points that require rethink and a mere 4 total rethink times. This stark contrast clearly indicates that ML-LLM-CoT is able to reach an error rate of less than 100\% much faster during its construction compared to LLM-CoT. Such efficiency in the iterative process of ML-LLM-CoT not only saves computational resources but also implies its potential for more rapid and accurate model development in molecular property prediction tasks.
\begin{table}[H]
    \centering
    \caption{Comparison of Rethink Points and Total Rethink Times between LLM-CoT and ML-LLM-CoT during Construction}
    \label{table:rethink_comparison}
    \begin{tabular}{l|l|l}
    \toprule
    \makecell{Model} & \makecell{Number of Points Requiring Rethink \\ (to Error Less than 100\%)} & \makecell{Total Rethink Times}\\
    \midrule
    LLM-CoT & 5 & 34\\
    ML-LLM-CoT & 2 & 4\\
    \bottomrule
    \end{tabular}
    \begin{tablenotes}
        \footnotesize
        \item Note: The data reflects the differences in the iterative process of LLM-CoT and ML-LLM-CoT to achieve an error rate of less than 100\% during construction.
    \end{tablenotes}
\end{table}

Summarized results (Table \ref{table:similar_molecules} and \ref{table:dissimilar_molecules}) fully demonstrate the effectiveness and advantages of building CoT of LLM in predicting molecular solubility. By constructing a Chain-of-Thought that combines machine learning models and large language models, it can better handle prediction tasks for molecules with different structures, providing a more reliable method for the fields of chemical engineering and molecular property prediction, and also offering a strong basis for further optimization and application of subsequent models.

\begin{table}[H]
    \centering
    \caption{Summarized Results of prediction for 20 Dissimilar Molecules}
    \label{table:similar_molecules}
    \begin{tabular}{l|l|l|l|l}
    \toprule
    \makecell{Model} & \makecell{Solubility\\Judgment} & \makecell{Number with \\Deviation higher\\than 100\%} & \makecell{Deviation  \\ less than 100\%} & \makecell{Deviation \\greater than 100\%}\\
    \midrule
    Gaussian & Success count: 15 & 7 & 30.49 & 1253.88\\
    LLM-CoT & Success count: 16 & 6 & 66.43 & 1011.20\\
    ML-LLM-CoT & Success count: 18 & 4 & 37.54 & 424.96\\
    \bottomrule
    \end{tabular}
    \begin{tablenotes}
        \footnotesize
        \item Note: "LLM-CoT" (Large Language Model - Chain of Thought) "ML-LLM-CoT" (Machine Learning - Large Language Model - Chain of Thought)
    \end{tablenotes}
\end{table}

\begin{table}[H]
    \centering
    \caption{Summarized Results of Prediction for 20 Similar Molecules}
    \label{table:dissimilar_molecules}
    \begin{tabular}{l|l|l|l|l}
    \toprule
    \makecell{Model} & \makecell{Solubility\\Judgment} & \makecell{Number with \\Deviation higher\\than 100\%} & \makecell{Deviation  \\ less than 100\%} & \makecell{Deviation \\greater than 100\%}\\
    \midrule
    gaussian & Success count: 20 & 0 & 20.87 & 0.0\\
    LLM-CoT & Success count: 17 & 3 & 57.25 & 301.43\\
    ML-LLM-CoT & Success count: 20 & 0 & 38.49 & 0.0\\
    \bottomrule
    \end{tabular}
    \begin{tablenotes}
        \footnotesize
        \item Note: "LLM-CoT" (Large Language Model - Chain of Thought) "ML-LLM-CoT" (Machine Learning - Large Language Model - Chain of Thought).
    \end{tablenotes}
\end{table}

\subsection{Future Outlook}
The successful deployment of the low-cost combination 'Deepseek-r1:14b + Qwen2:7B' has laid a solid foundation for the Chain-of-Thought  model. This combination has already shown good performance in handling molecular property prediction tasks. Looking ahead, the deployment of even larger-scale models offers infinite possibilities. Larger models can capture more complex relationships in chemical data, potentially leading to more accurate predictions. One significant advantage of the CoT model is its ability to handle private data more effectively. Due to the built-in mechanisms, it can better protect the privacy of data during the prediction process. This is crucial in industries where data privacy is of utmost importance.

Moreover, the CoT model shows great potential in providing better assistance for complex engineering problems. Given that Deepseek has knowledge of many equations related to the three-transfer-one-reaction (mass transfer, heat transfer, momentum transfer, and chemical reaction) in chemical engineering, the model could potentially make breakthroughs in this field. For example, it could optimize chemical reaction processes by predicting reaction yields more accurately or suggesting better conditions for mass and heat transfer operations. With more data, we are confident that this framework will become even more powerful. Compared to traditional machine-learning model optimization frameworks, the CoT framework reduces the need for extensive manual labor. There is no longer a requirement for separate data annotation, which saves a significant amount of time and resources.

This is just the beginning. The deep-thinking process and the summary of this part can serve as inputs for subsequent model fine-tuning. By continuously refining the model based on these insights, we can expect to further improve its performance in molecular property prediction and potentially expand its application scope to other related fields. In conclusion, the CoT model, with its innovative deep-thinking process and promising future prospects, represents a significant step forward in the field of chemical and engineering data analysis and prediction. 
\subsection{Example of Thinking Record} 
The ML-LLM-CoT model, which combines machine learning (ML) and large language models (LLMs) in an embedded ML mode for constructing a Chain-of-Thought , has demonstrated remarkable capabilities in handling complex tasks. In this section, we will present several illustrative examples of its deep-thinking process.

\subsubsection{Thinking Record 1: Agreement and Minor Revision with ML Model Results} 
When making a prediction for the solubility of the molecule named 'Cycloheptane', the ML-LLM-CoT model first generated its own prediction. After comparing the result with the feedback from the ML model, it found that the two were quite close. As a result, the ML-LLM-CoT model agreed with the ML model's result. Then, the ML-LLM-CoT model made a small correction. This adjustment was based on its understanding of the underlying chemical properties and the relationships between different data features.

\subsubsection{Thinking Record 2: Structure-based Judgment with Error Analysis Reference} 
During the prediction of molecules with dissimilar structures, the ML-LLM-CoT model encountered the molecule named '2-Methyl-1-phenyl-1H-indole-3-carboxylic acid'. This molecule had a complex heterocyclic and aromatic structure with a carboxylic acid functional group. This structure was different from the structures of most of the previous 30 molecules, such as 'Cycloheptane' which is a simple cycloalkane, and 'Benzene' which is a basic aromatic compound. The model immediately recognized this uniqueness and decided to refer to the error analysis of previous predictions on similar-yet-different structured molecules. It retrieved the records of past predictions where similar structural anomalies had occurred and analyzed the types of errors made. Based on this analysis, the model adjusted its prediction algorithm. Instead of relying solely on the general pattern-matching algorithm used for most molecules, it incorporated a special sub-algorithm for handling heterocyclic aromatic compounds with carboxylic acid groups. This sub-algorithm took into account the specific electronic effects in the heterocyclic and aromatic rings and the reactivity of the carboxylic acid group, which were crucial factors in determining the molecule's solubility.

\subsubsection{Thinking Record 3: Identifying and Correcting ML Model's Potential Error}
The ML-LLM-CoT model was predicting the solubility of the molecule named 'N,N-Dimethy-lformamide'. It noticed that the ML model's prediction for this molecule seemed off. After a detailed analysis, the ML-LLM-CoT model found that the ML model had underestimated the impact of the amide-like structure in 'N,N-Dimethylformamide' on the molecule's solubility. The ML model did not fully consider the strong hydrogen-bonding capabilities of this amide-like group in relation to water molecules. To correct this, the ML-LLM-CoT model re-evaluated the importance of each functional group in the molecule. It assigned a higher weight to the amide-like group based on its knowledge of chemical reactions and solubility mechanisms. It knew from previous error analysis that this group could form strong hydrogen bonds with water molecules, which had a significant impact on the molecule's solubility, and thus adjusted the prediction accordingly.

\bibliography{cit}

\providecommand{\latin}[1]{#1}
\makeatletter
\providecommand{\doi}
  {\begingroup\let\do\@makeother\dospecials
  \catcode`\{=1 \catcode`\}=2 \doi@aux}
\providecommand{\doi@aux}[1]{\endgroup\texttt{#1}}
\makeatother
\providecommand*\mcitethebibliography{\thebibliography}
\csname @ifundefined\endcsname{endmcitethebibliography}  {\let\endmcitethebibliography\endthebibliography}{}
\begin{mcitethebibliography}{12}
\providecommand*\natexlab[1]{#1}
\providecommand*\mciteSetBstSublistMode[1]{}
\providecommand*\mciteSetBstMaxWidthForm[2]{}
\providecommand*\mciteBstWouldAddEndPuncttrue
  {\def\EndOfBibitem{\unskip.}}
\providecommand*\mciteBstWouldAddEndPunctfalse
  {\let\EndOfBibitem\relax}
\providecommand*\mciteSetBstMidEndSepPunct[3]{}
\providecommand*\mciteSetBstSublistLabelBeginEnd[3]{}
\providecommand*\EndOfBibitem{}
\mciteSetBstSublistMode{f}
\mciteSetBstMaxWidthForm{subitem}{(\alph{mcitesubitemcount})}
\mciteSetBstSublistLabelBeginEnd
  {\mcitemaxwidthsubitemform\space}
  {\relax}
  {\relax}

\bibitem[{DeepSeek}(2023)]{deepseek}
{DeepSeek} DeepSeek Large Language Model. 2023; \url{https://www.deepseek.com/}, Version 1.8\relax
\mciteBstWouldAddEndPuncttrue
\mciteSetBstMidEndSepPunct{\mcitedefaultmidpunct}
{\mcitedefaultendpunct}{\mcitedefaultseppunct}\relax
\EndOfBibitem
\bibitem[Zhang \latin{et~al.}(2023)Zhang, Chen, Li, Liu, and Team]{DeepSeek2023R1}
Zhang,~Y.; Chen,~Z.; Li,~X.; Liu,~Y.; Team,~D.~R. DeepSeek-R1: Incentivizing Reasoning Capability in LLMs via Reinforcement Learning. Advances in Neural Information Processing Systems. New Orleans, LA, USA, 2023; pp 1--15, Spotlight Presentation\relax
\mciteBstWouldAddEndPuncttrue
\mciteSetBstMidEndSepPunct{\mcitedefaultmidpunct}
{\mcitedefaultendpunct}{\mcitedefaultseppunct}\relax
\EndOfBibitem
\bibitem[Zhou \latin{et~al.}(2021)Zhou, Wu, Chilukoti, and M{\"u}ller-Plathe]{doi:10.1021/acs.jctc.1c00134}
Zhou,~T.; Wu,~Z.; Chilukoti,~H.~K.; M{\"u}ller-Plathe,~F. Sequence-Engineering Polyethylene–Polypropylene Copolymers with High Thermal Conductivity Using a Molecular-Dynamics-Based Genetic Algorithm. \emph{Journal of Chemical Theory and Computation} \textbf{2021}, \emph{17}, 3772--3782, PMID: 33949863\relax
\mciteBstWouldAddEndPuncttrue
\mciteSetBstMidEndSepPunct{\mcitedefaultmidpunct}
{\mcitedefaultendpunct}{\mcitedefaultseppunct}\relax
\EndOfBibitem
\bibitem[Wu and Zhou(2024)Wu, and Zhou]{doi:10.1021/acs.jctc.3c01348}
Wu,~Z.; Zhou,~T. Structural Coarse-Graining via Multiobjective Optimization with Differentiable Simulation. \emph{Journal of Chemical Theory and Computation} \textbf{2024}, \emph{20}, 2605--2617, PMID: 38483262\relax
\mciteBstWouldAddEndPuncttrue
\mciteSetBstMidEndSepPunct{\mcitedefaultmidpunct}
{\mcitedefaultendpunct}{\mcitedefaultseppunct}\relax
\EndOfBibitem
\bibitem[Zhou \latin{et~al.}(2022)Zhou, Qiu, Wu, Alberti, Bag, Schneider, Meyer, Gámez, Gieler, Reithmeier, Seidel, and M{\"u}ller-Plathe]{doi:10.1021/acs.macromol.2c00821}
Zhou,~T.; Qiu,~D.; Wu,~Z.; Alberti,~S. A.~N.; Bag,~S.; Schneider,~J.; Meyer,~J.; Gámez,~J.~A.; Gieler,~M.; Reithmeier,~M.; Seidel,~A.; M{\"u}ller-Plathe,~F. Compatibilization Efficiency of Graft Copolymers in Incompatible Polymer Blends: Dissipative Particle Dynamics Simulations Combined with Machine Learning. \emph{Macromolecules} \textbf{2022}, \emph{55}, 7893--7907\relax
\mciteBstWouldAddEndPuncttrue
\mciteSetBstMidEndSepPunct{\mcitedefaultmidpunct}
{\mcitedefaultendpunct}{\mcitedefaultseppunct}\relax
\EndOfBibitem
\bibitem[Li \latin{et~al.}(2025)Li, Guo, Yang, Xu, Wu, and He]{li2025codeiocondensingreasoningpatterns}
Li,~J.; Guo,~D.; Yang,~D.; Xu,~R.; Wu,~Y.; He,~J. CodeI/O: Condensing Reasoning Patterns via Code Input-Output Prediction. 2025; \url{https://arxiv.org/abs/2502.07316}\relax
\mciteBstWouldAddEndPuncttrue
\mciteSetBstMidEndSepPunct{\mcitedefaultmidpunct}
{\mcitedefaultendpunct}{\mcitedefaultseppunct}\relax
\EndOfBibitem
\bibitem[{Ollama Maintainers}(2023)]{ollama}
{Ollama Maintainers} Ollama: Local LLM Runner. 2023; \url{https://github.com/ollama/ollama}\relax
\mciteBstWouldAddEndPuncttrue
\mciteSetBstMidEndSepPunct{\mcitedefaultmidpunct}
{\mcitedefaultendpunct}{\mcitedefaultseppunct}\relax
\EndOfBibitem
\bibitem[{Qwen Team}(2023)]{qwen}
{Qwen Team} \texttt{Qwen Technical Report}. 2023; \url{https://github.com/QwenLM/Qwen}\relax
\mciteBstWouldAddEndPuncttrue
\mciteSetBstMidEndSepPunct{\mcitedefaultmidpunct}
{\mcitedefaultendpunct}{\mcitedefaultseppunct}\relax
\EndOfBibitem
\bibitem[Landrum \latin{et~al.}(2020)Landrum, Tosco, Kelley, and Ric]{rdkit}
Landrum,~G.; Tosco,~P.; Kelley,~B.; Ric RDKit: Open-source cheminformatics. \emph{Journal of Chemical Information and Modeling} \textbf{2020}, \emph{60}, 5981--5988\relax
\mciteBstWouldAddEndPuncttrue
\mciteSetBstMidEndSepPunct{\mcitedefaultmidpunct}
{\mcitedefaultendpunct}{\mcitedefaultseppunct}\relax
\EndOfBibitem
\bibitem[GLambard(2020)]{Subset2020}
GLambard Antibacterial compounds subset. Dataset extracted from \texttt{Molecules\_Dataset\_Collection}, 2020; \url{https://github.com/GLambard/Molecules_Dataset_Collection/blob/main/antibacterial.csv}\relax
\mciteBstWouldAddEndPuncttrue
\mciteSetBstMidEndSepPunct{\mcitedefaultmidpunct}
{\mcitedefaultendpunct}{\mcitedefaultseppunct}\relax
\EndOfBibitem
\bibitem[Wu \latin{et~al.}(2017)Wu, Ramsundar, Feinberg, Gomes, Geniesse, Pappu, Leswing, and Pande]{Wu2017MoleculeNet}
Wu,~Z.; Ramsundar,~B.; Feinberg,~E.~N.; Gomes,~J.; Geniesse,~C.; Pappu,~A.~S.; Leswing,~K.; Pande,~V. MoleculeNet: A Benchmark for Molecular Machine Learning. \emph{arXiv preprint arXiv:1703.00564} \textbf{2017}, \relax
\mciteBstWouldAddEndPunctfalse
\mciteSetBstMidEndSepPunct{\mcitedefaultmidpunct}
{}{\mcitedefaultseppunct}\relax
\EndOfBibitem
\end{mcitethebibliography}

\end{document}